\documentclass{article}
\usepackage{ijcai25}

\usepackage{times}
\usepackage{soul}
\usepackage{url}
\usepackage[hidelinks]{hyperref}
\usepackage[utf8]{inputenc}
\usepackage[small]{caption}
\usepackage{graphicx}
\usepackage{amsmath}
\usepackage{amsthm}
\usepackage{booktabs}
\usepackage{algorithm}
\usepackage{algorithmic}
\usepackage[switch]{lineno}

\usepackage{xcolor}
\usepackage{multirow}
\usepackage{stfloats}
\usepackage{amsfonts}
\usepackage[caption=false,font=small,labelfont=rm,textfont=rm]{subfig}
\usepackage[normalem]{ulem}
\usepackage[misc]{ifsym}

\title{Are KANs Effective for Multivariate Time Series Forecasting?}

\author{
Xiao Han$^{1,2}$\and
Xinfeng Zhang$^{1,}$\textsuperscript{\Letter}\and
Yiling Wu$^{2,}$\textsuperscript{\Letter}\and
Zhenduo Zhang$^{1,2}$\And
Zhe Wu$^{2}$\\
\affiliations
$^{1}$School of Computer Science and Technology, \\University of Chinese Academy of Sciences, Beijing, China \\
$^{2}$Pengcheng Laboratory, Shenzhen, China \\
\emails
\{hanxiao22,zhangzhenduo21\}@mails.ucas.ac.cn,
xfzhang@ucas.ac.cn,
\{wuyl02,wuzh02\}@pcl.ac.cn
}

\begin{document}
\maketitle
\begin{abstract}
Multivariate time series forecasting is a crucial task that predicts the future states based on historical inputs. Related techniques have been developing in parallel with the machine learning community, from early statistical learning methods to current deep learning methods. Despite their significant advancements, existing methods continue to struggle with the challenge of inadequate interpretability. The rise of the Kolmogorov-Arnold Network (KAN) provides a new perspective to solve this challenge, but current work has not yet concluded whether KAN is effective in time series forecasting tasks. In this paper, we aim to evaluate the effectiveness of KANs in time-series forecasting from the perspectives of performance, integrability, efficiency, and interpretability. To this end, we propose the Multi-layer Mixture-of-KAN network (MMK), which achieves excellent performance while retaining KAN's ability to be transformed into a combination of symbolic functions. The core module of MMK is the mixture-of-KAN layer, which uses a mixture-of-experts structure to assign variables to best-matched KAN experts. Then, we explore some useful experimental strategies to deal with the issues in the training stage. Finally, we compare MMK and various baselines on seven datasets. Extensive experimental and visualization results demonstrate that KANs are effective in multivariate time series forecasting. Code is available at: https://github.com/2448845600/EasyTSF.
\end{abstract}

\section{Introduction}

\begin{figure}[t]
\centering
\subfloat[Performance comparison]{\includegraphics[width=0.22\textwidth]{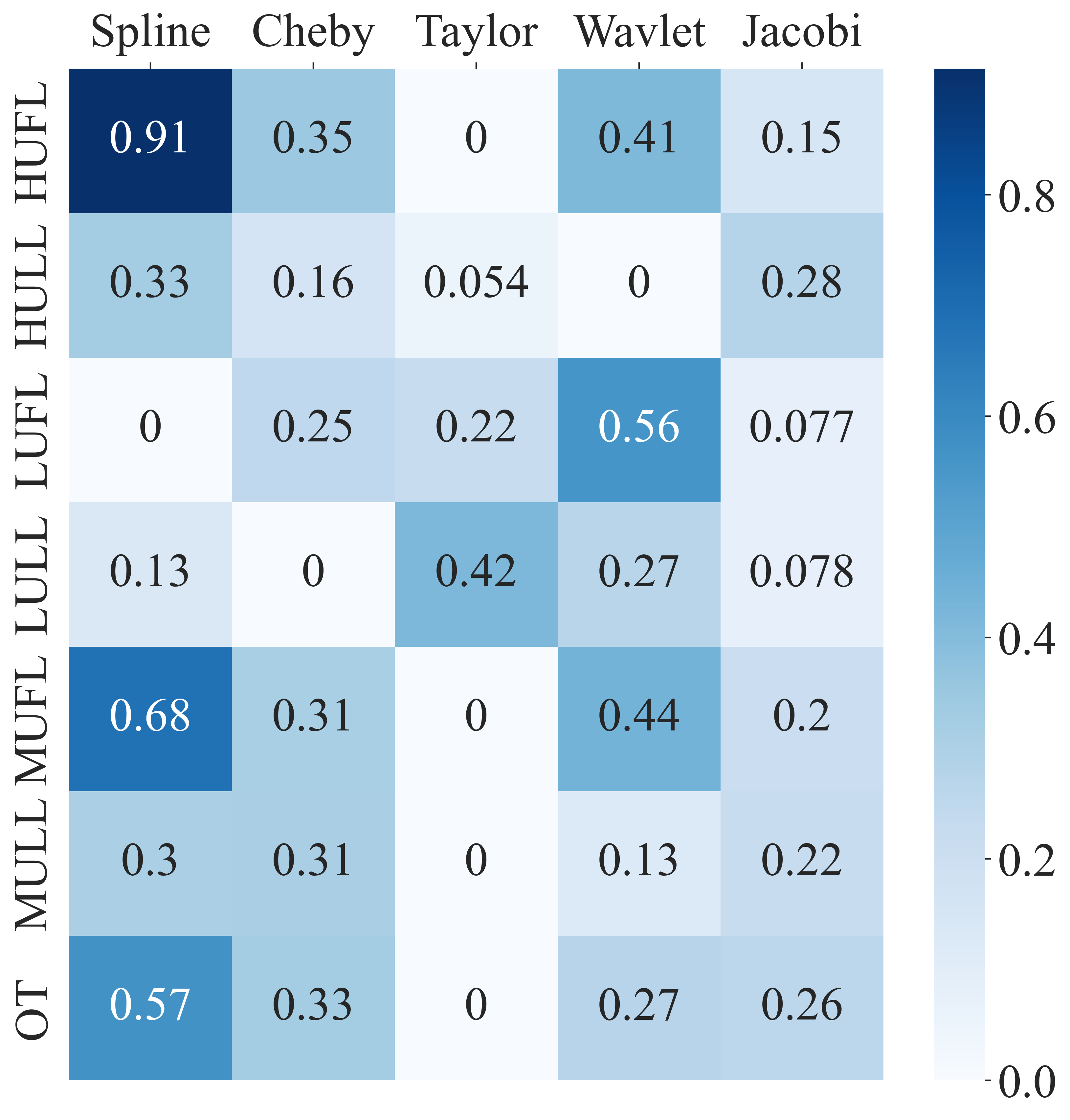}}
\subfloat[Training Loss]{\includegraphics[width=0.22\textwidth]{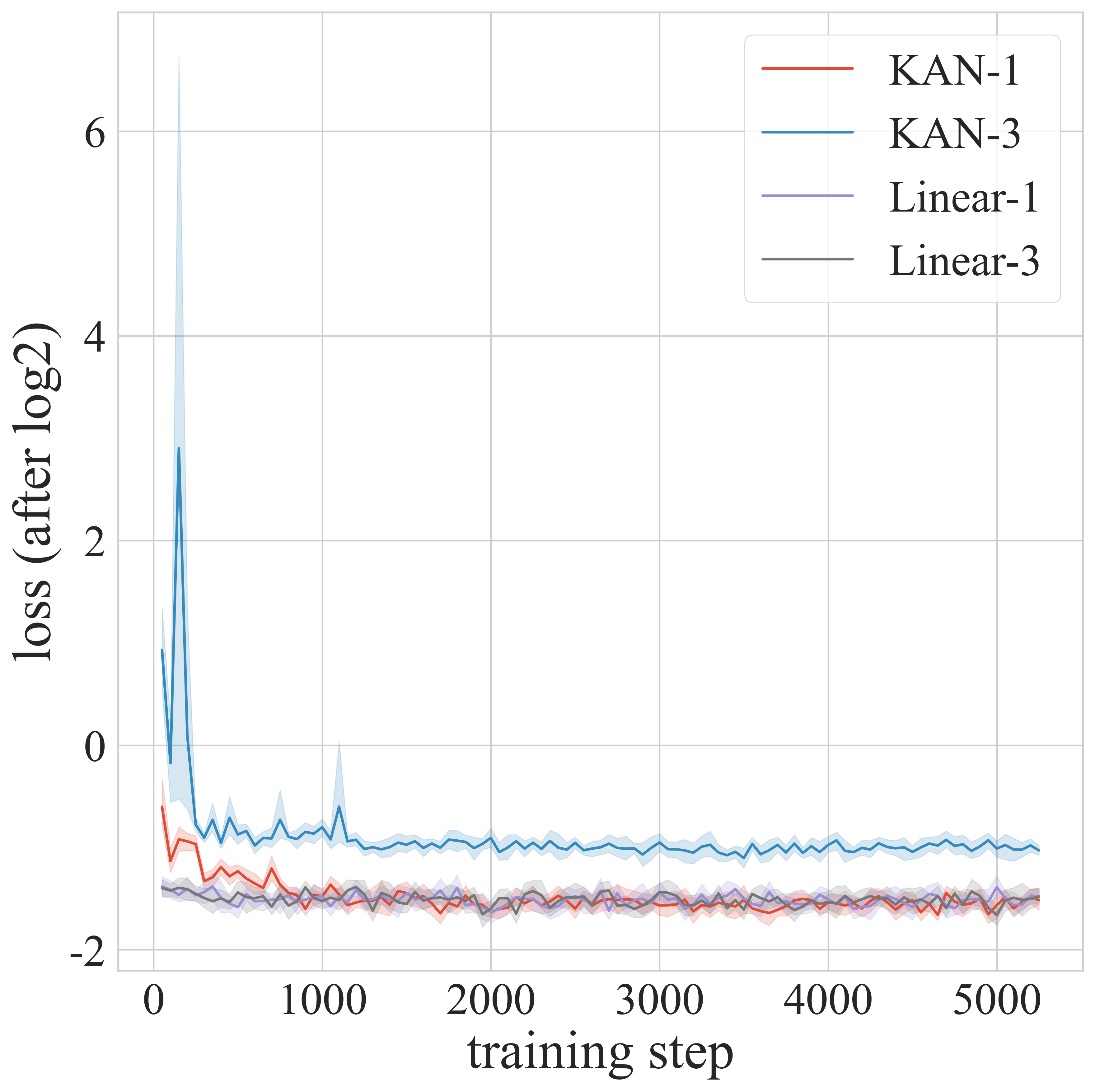}}
\caption{Two experiental findings on the ETTh1 dataset. Subfigure (a) shows the univariate forecasting results of KANs with different base functions, where results are minmax-normalized and 0 means the minimum MSE in the row. Subfigure (b) reports the training loss of one/three layers of Linear/KAN models.}
\label{fig:experiental-findings}
\end{figure}

The multivariate time series forecasting (MTSF) task aims to predict future states of variables from input historical data. Related technologies have been applied to a broad scope of fields, such as financial investment, weather forecasting, and traffic management \cite{gao2023stockformer,savcisens2024using,han2024etf}. Time series forecasting methods have long been inspired by advances in the machine learning community. The popularity of early statistical learning methods gave rise to SVR and ARMIA \cite{holt2004forecasting}; recurrent neural networks (RNN), which are specially designed to process sequence data, are naturally applied to time series analysis \cite{hewamalage2021recurrent}; TFT \cite{lim2021tft} and Informer \cite{zhou2021informer} promote the application of transformer architecture in time series prediction; LSTF-Linear \cite{zeng2023dlinear} rethinks the value of Linear. At present, various deep learning network architectures, such as RNN, CNN, Transformer, and MLP, are widely used in time series forecasting \cite{wang2023micn,zhou2024sdformer,jia2024witran}. The forecasting models derived from different network architectures have their own advantages in forecasting performance, running speed, and resource usage.

However, most popular time series forecasting models are based on deep learning-based modules, whose prediction process is blackbox and prediction results lack interpretability. These nontransparent methods are suspected to be unsuitable for tasks with low error tolerance, such as medicine, law, and finance. Some traditional, non-deep learning-based methods are interpretable, but it is difficult to achieve the expected performance. Fortunately, a novel kind of deep learning network, the Kolmogorov-Arnold Network (KAN) \cite{liu2024kan}, which is based on the Kolmogorov-Arnold representation theorem (KART), has become a feasible solution to the above dilemma of choosing between performance and interpretability. KAN offers a pruning strategy that simplifies the trained KAN model into a set of symbolic functions, making it possible to analyze the mechanisms of the network, thereby significantly enhancing the interpretability. 

Although KAN has the advantages of both performance and interpretability in theory, some works \cite{yu2024kan} have pointed out that it does not work well for real-world tasks. A few works \cite{genet2024temporal,vaca2024kolmogorov} have tried to apply KAN to time series analysis, but they have not used the characteristics of time series data to design efficient KAN-based networks, and lack fair comparisons on common datasets. Existing works cannot provide a convincing conclusion on whether KAN is effective in time series forecasting. Therefore, we aim to provide a reliable benchmark to comprehensively evaluate the effectiveness of the KAN in multivariate time series forecasting tasks for the first time. When applying KAN to time series forecasting, we encountered two challenges, as following. 

One is \textbf{how to select the most suitable KAN}. The KAN proposed by Ziming Liu et al. is not only a model but also a general function-fitting framework. Subsequent studies \cite{ss2024chebyshev,bozorgasl2024wav,aghaei2024fkan} use different base functions to replace the B-spline function for better performance on specific tasks. We report the experimental results about five KAN's variants (KANs), each with a special base function, on seven univariates of the ETTh1 dataset, as illustrated in Figure \ref{fig:experiental-findings}(a). This makes it cumbersome to determine the appropriate KAN for achieving the best performance. The other is \textbf{how to train the deeper KAN}. We build a single-layer KAN model that can achieve similar performance with a single-layer Linear model. But as the layer of the KAN model increases, the performance drops significantly. We visualize the values of loss in 10 times repeated trainings in Figure \ref{fig:experiental-findings}(b) and find that the initial loss of KAN is larger than that of Linear, and the loss of the 3-layer KAN model becomes significantly larger and more unstable.

To address the above challenges, we first apply the mixture-of-experts technique to KAN and obtain the Mixture-of-KAN (MoK) layer, which uses a gating network to adaptively assign the input to the matching KAN layers with special base functions according to the variable dimensions. This structure aggregates the advantages of KAN's variants in different cases. Then, we build a simple yet efficient Multi-layer Mixture-of-KAN network (MMK) based on MoK. By using modules that do not compromise the ability to be converted into symbolic functions, MMK achieves excellent performance while retaining the interpretability from KAN. Next, we analyze the phenomenon of abnormal training loss and use some training strategies to improve the performance of the proposed MMK. Finally, we evaluate the effectiveness of KANs in multivariate time series forecasting from four perspectives: performance, integration, speed, and interpretability. We compare MMK with many popular time series forecasting models on seven commonly used datasets, and experimental results show that MMK achieves state-of-the-art performance in most cases. We integrate MoK into recently proposed models and find that it can improve their performance. And we visualize the output weights of the gating network in the MoK layer and analyze the connection between KAN's symbolic function conversion ability and domain knowledge.

To sum up, the contributions of this paper include:
\begin{itemize}
    \item To the best of our knowledge, this is the first work that comprehensively discusses the effectiveness of the booming KANs for time series forecasting.
    \item To validate our claims, we propose an interpretable Multi-layer Mixture-of-KAN network (MMK), which uses the mixture-of-KAN layer to aggregate the advantages of KAN's variants. And we use some training strategies to improve performance.
    \item We fairly compare MMK with many baselines on seven real-world datasets, and the experimental results show that MMK achieves the competitive performance.
\end{itemize}

In summary, we can conclude that \textbf{\uline{KANs are effective in multivariate time series forecasting}}.

\section{Related Work}

\subsection{Multivariate Time-Series Forecasting}
Various network architectures (such as Transformer, CNN, and MLP) are competing in time series forecasting recently. 
Transformer-based time series forecasting models have strong performance but high time and memory complexity. Informer \cite{zhou2021informer} proposed ProbSparse self-attention to reduce the complexity from $\mathcal{O}(T^2)$ to $\mathcal{O}(T \log T)$. Pyraformer \cite{Liu2022PyraformerLP} utilizes the pyramid attention mechanism to capture hierarchical multi-scale time dependencies with a time and memory complexity of $\mathcal{O}(T)$. PatchTST \cite{nie2023patchtst} and Crossformer \cite{zhang2023crossformer} use the patch operation to reduce the number of input tokens, thereby reducing time complexity. 
The performance of early Multi-layer perception-based (MLP-based) models is generally weaker than transformer-based methods. However, NLinear \cite{zeng2023dlinear} and RLinear \cite{Li2023RevisitingLT} combine different normalization methods with a single-layer MLP, and achieve performance that exceeds Transformer-based models on some datasets with extremely low computational cost. Recurrent neural networks (RNNs) are suitable for handling sequence data, making them a favorable choice for time series analysis. The size of the hidden state in RNN is independent of the input time series length, so recent RNN-based time series prediction models, such as SegRNN \cite{lin2023segrnn} and WITRAN \cite{jia2024witran}, apply the RNN structure to time series forecasting with longer input. Convolutional neural networks (CNNs) are frequently used in time series forecasting models in the form of 1D convolution, such as ModernTCN \cite{donghao2024moderntcn} and SCINet \cite{liu2022SCINet}, but TimesNet \cite{wu2023timesnet} takes a different approach by converting the 1D time series into a 2D matrix through the Fourier transform and then using 2D convolution for prediction. 

\subsection{Kolmogorov-Arnold Network}

The Kolmogorov-Arnold representation theorem (KART) is the mathematical foundation of the Kolmogorov-Arnold Network (KAN) \cite{liu2024kan}, which makes KAN more fitting and interpretable than Multi-Layer Perceptrons (MLP) based on the universal approximation theorem.
Given an input tensor $\mathbf{x} \in \mathbb{R}^{n_0}$, the structure of $L$ layers in a KAN network can be represented as:
\begin{equation}
    \operatorname{KAN}(\mathbf{x}) = \left(\boldsymbol{\Phi}_{L} \circ \boldsymbol{\Phi}_{L-1} \circ \cdots \circ \boldsymbol{\Phi}_{2} \circ \boldsymbol{\Phi}_{1}\right) \mathbf{x},
\end{equation}
where $\boldsymbol{\Phi}_{l}, l \in [1, 2, \cdots, L]$ is a KAN layer, and the output dimension of each KAN layer can be expressed as: $[n_1, n_2, \cdots, n_L]$. Therefore, the transform process of $j$-th feature in $l$-th layer can be formed as:
\begin{equation}
\mathbf{x}_{l, j}=\sum_{i=1}^{n_{l-1}} \phi_{l-1, j, i}\left(\mathbf{x}_{l-1, i}\right), \quad j=1, \cdots, n_{l},
\end{equation}
where $\phi$ consists two parts: the spline function and the residual activation function with learnable parameters $w_b, w_s$:
\begin{equation}
    \phi(x) = w_a \mathrm{SiLU}(x) + w_b\mathrm{Spline}(x),
\label{eq:phi}
\end{equation}
where $\mathrm{Spline}(\cdot)$ is a linear combination of B-spline functions $\mathrm{Spline}(x) = \sum_i c_{i}B_{i}(x)$. 

Recent studies have proposed many variants of KAN to expand its application scope and improve its performance. Some work uses functions such as wavelet functions \cite{bozorgasl2024wav}, Taylor polynomials \cite{yu2024taylorkan} and Jacobi polynomials \cite{aghaei2024fkan} instead of spline functions to improve the performance of KAN in different fields. We can unify these KANs using the generalized version of Equation \ref{eq:phi} as: 
\begin{equation}
    \phi(x) = w_a \mathbb{A}(x) + w_b\mathbb{F}(x).
\end{equation}


\section{Multi-layer Mixture-of-KAN Network}

\subsection{Problem Defintion}
In multivariate time series forecasting, given historical data $\mathcal{X} = [X_1, \cdots, X_T] \in \mathbb{R}^{T \times C}$, where $T$ is the time steps of historical data and $C$ is the number of variates. The time series forecasting task is to predict $\mathcal{Y} = [X_{T+1}, \cdots, X_{T+P}] \in \mathbb{R}^{P \times C}$ during future $P$ time steps.

\subsection{Mixture-of-KAN Layer}
Real-world time series data frequently exhibit non-stationarity, with their statistical properties (such as mean and variance) varying over time. Moreover, there are significant distribution discrepancies between variables in multivariate time series. This poses a significant challenge to time series forecasting techniques, inevitably impacting the KAN-based methods as well. Fortunately, when it comes to dealing with distribution shift, KAN has a unique characteristic compared to existing methods: it has many variants using different base functions. Considering that the special base function may be suitable for modeling certain data distributions, we try to combine several KANs into a single layer and adaptively schedule them according to the input data.

Following this idea, we propose the Mixture-of-KAN (MoK) layer, which is a combination of KAN and mixture-of-experts (MoE). The MoK layer uses a gating network to assign KAN layers to variables according to temporal features, where each expert is responsible for a specific part of the data. KAN and its variants only differ in the base function, so we use $\mathcal{K}(\cdot)$ to represent KANs uniformly in this paper. Our proposed MoK layer with $N$ experts can be simply formed as: 
\begin{equation}
    \mathbf{x}_{l+1} = \sum_{i=1}^{N} \mathcal{G}(\mathbf{x}_{l})_{i}\mathcal{K}_{i}(\mathbf{x}_{l}),
\end{equation}
where $\mathcal{G}(\cdot)$ is a gating network. 
This mixture of expert structure can adapt to the diversity of time series, with each expert learning different parts of temporal features, thus improving performance on time series forecasting task.

The gating network is the key module of the MoK layer, which is responsible for learning the weight of each expert from the input data. The softmax gating network $\mathcal{G}_{\mathrm{softmax}}$ which uses a softmax function and a learnable weight matrix $\mathbf{w}_g$ to schedule input data as:
\begin{align}
\mathcal{G}_{\mathrm{softmax}}(\mathbf{x}) = \mathrm{Softmax}(\mathbf{x} \mathbf{w}_g).
\end{align}
This gating network is popular due to its simple structure. However, it activates all experts once, resulting in low efficiency when there are a large number of experts. Therefore, we adopt the sparse gating network \cite{shazeer2017outrageously} which only activates the best matching top-$k$ experts. It adds Gaussian noise to input time series by $\mathbf{w}_{\mathrm{noise}}$, and uses $\mathrm{KeepTopK}$ operation to retain experts with the highest $k$ values. The sparse gating network can be described as:
\begin{align}
\mathcal{G}_{\mathrm{sparse}}(\mathbf{x}) = \mathrm{Softmax}(\mathrm{KeepTopK}(H(\mathbf{x}), k)), \\
H(\mathbf{x}) = \mathbf{x} \mathbf{w}_g + \mathrm{Norm}(\mathrm{Softplus}(\mathbf{x} \mathbf{w}_{\mathrm{noise}})),
\end{align}
where $\mathrm{Norm}(\cdot)$ is standard normalization.

\begin{figure}[t]
\centering
\includegraphics[width=0.48\textwidth]{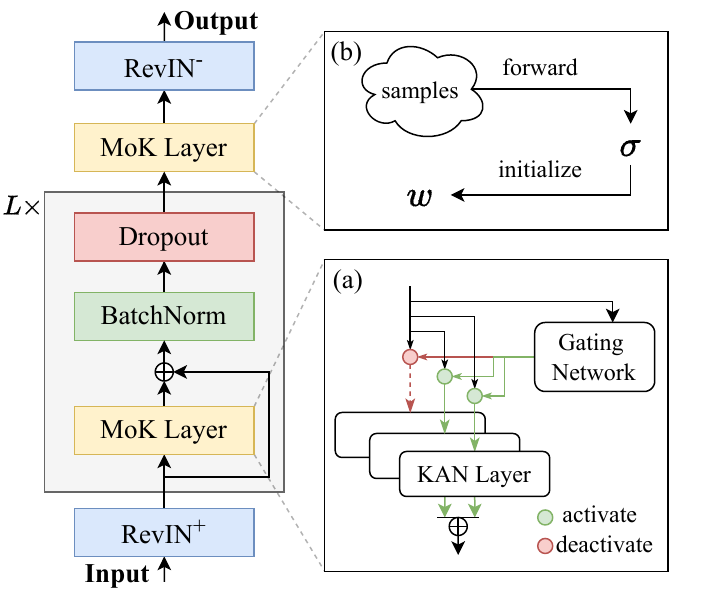}
\caption{The architecture of MMK. Subfigure (a) is the structure of the mixture-of-KAN (MoK) layer; subfigure (b) is the pre-sample initialization strategy for the last KAN layer.}
\label{fig:pipeline}
\end{figure}

\subsection{Network Architecture}
The interpretability of KAN comes from the fact that the trained KAN model can be converted into a combination of many simple symbolic functions. In order to maintain this advantage, we use layers or operators with the same properties to build our KAN-based forecasting model.

\paragraph{RevIN} RevIN \cite{kim2022reversible} includes symmetrically structured normalization $\text{RevIN}^{+}$ and denormalization operations $\text{RevIN}^{-}$ in the variable dimension. Before sending data $X$ to the backbone network, $\text{RevIN}^{+}$ uses a learnable affine transformation to normalize the input time series of each variable. After the last MoK layer, outputs are denormalized to the original distribution space using the same affine transformation parameters in the first step by $\text{RevIN}^{-}$ operation. Each learnable parameter of RevIN corresponds to the statistical value of the time series variable, and it can be viewed as a symbolic function in form, so it does not harm the interpretability of the overall network.


\paragraph{MoK Block} Inspired by the ResNet \cite{he2016deep}, we design a simple MoKBlock, which consists of a MoK layer, BatchNorm, and Dropout in sequence, and adds a residual structure. By stacking several MoKBlocks, MMK can obtain the stronger feature extraction capability.

\begin{equation}
\operatorname{MoKBlock}(\boldsymbol{h}) = \operatorname{Dropout}(\operatorname{BN}(\boldsymbol{h} + \operatorname{MoK}(\boldsymbol{h}))).
\end{equation}

\paragraph{Loss Function} During the training stage, the gating network has a tendency to reach a winner-take-all state where it always gives large weights for the same few experts. Following the previous work \cite{shazeer2017outrageously}, we add a load balancing loss function to encourage experts have equal importance. First, we count the weight of experts as loads, and calculate the square of the coefficient of variation ($\operatorname{CV}(\cdot)$) of the load values as additional loss:
\begin{equation}
    \mathcal{L}_{\text{load-balancing}} = \operatorname{CV}(\text{loads})^{2}.
\end{equation}
With this loss function, all experts are encouraged to have equal importance in the training stage. 
The total loss function is the sum of the prediction loss and load balancing loss:
\begin{equation}
\begin{aligned}
\mathcal{L} = \mathcal{L}_{\text{pred}} + \mathcal{L}_{\text{load-balancing}}, \\
\text{where } \mathcal{L}_{\text{pred}} = \operatorname{MSE}(\mathcal{Y}, \hat{\mathcal{Y}}).
\end{aligned}
\end{equation}

\subsection{Training Strategy}
Although our proposed MMK is carefully designed, the performance of multiple-layer MMK is not as good as that of a single-layer MMK. From the visualization of the training loss, we infer that the main issue in training deep MMK is that the initial loss is too large and unstable. In the original KAN paper, $w_a$ is initialized according to the Xavier initialization and $w_b$ is initialized to 1 to guarantee variance-preserving \cite{glorot2010understanding,he2015delving,yang2024kat} with the assumption that all $x_i$ in $x$ are mutually independent. However, this initialization method is unstable when the number of network layers increases, because its premise, the mathematical expectation remains 0, is heavily dependent on the distribution of input data. We compare expectations of KAN and Linear in Equation \ref{eq:expectations}:

\begin{equation}
\begin{aligned}
\mathbb{E}(\boldsymbol{y})^{\text{KAN}} &= \mathbb{E}[\phi(\boldsymbol{x})] \\
&= \mathbb{E}[\boldsymbol{w}_a]\mathbb{E}[\mathbb{A}(\boldsymbol{x})] + \mathbb{E}[\boldsymbol{w}_b]\mathbb{E}[\mathbb{F}(\boldsymbol{x})],\\
&= \mathbb{E}[\mathbb{F}(\boldsymbol{x})],\\
\mathbb{E}(\boldsymbol{y})^{\text{Linear}} & = \mathbb{E}[\boldsymbol{w}\boldsymbol{x}] = \mathbb{E}[\boldsymbol{w}]\mathbb{E}[\boldsymbol{x}] = 0.\\
\end{aligned}
\label{eq:expectations}
\end{equation}

The Linear layer keeps $\mathbb{E}=0$ by initializing $\boldsymbol{w}=0$, while KAN keeps $\mathbb{E}=0$ by supposing $\mathbb{E}[\mathbb{F}(\boldsymbol{x})]=0$. Since the real-world time series data has a biased distribution and non-negligible number of outliers, it is difficult to maintain $\mathbb{E}[\mathbb{F}(\boldsymbol{x})]=0$. Therefore, we propose a pre-sampling initialization strategy: before training starts, we forward all training samples to the model, and record the $\operatorname{Var}[\boldsymbol{x}]$ and $\operatorname{Var}[\mathbb{F}(\boldsymbol{x})]$ to initialize the $\boldsymbol{w}_b$ with $\mathcal{N}(0, \frac{\operatorname{Var}[\mathbb{F}(\boldsymbol{x})]}{\operatorname{Var}[\boldsymbol{x}]})$.

\begin{table}[]
\centering 
\begin{tabular}{cccc}
\midrule
Dataset      & Variates  & Timesteps & Granularity \\ 
\midrule
ETTh1/h2        & 7         & 17,420    & 1 hour \\ 
ETTm1/m2        & 7         & 69,680    & 15 min \\
ECL          & 321       & 26,304    & 1 hour \\
Traffic      & 862       & 17,544    & 1 hour \\
Weather      & 21        & 52,696    & 10 min \\
\bottomrule 
\end{tabular}
\caption{The statistics of the used datasets.}
\label{tab:dataset}
\end{table}

\begin{table*}[]
\small
\centering 
\setlength{\tabcolsep}{4pt}
\begin{tabular}{cc|cc|cc|cc|cc|cc|cc|cc|cc}
\toprule
\multicolumn{2}{c|}{}                         & \multicolumn{2}{c|}{MMK}                                     & \multicolumn{2}{c|}{iTransformer}                            & \multicolumn{2}{c|}{PatchTST}                                & \multicolumn{2}{c|}{FEDformer}        & \multicolumn{2}{c|}{TimesNet}                                & \multicolumn{2}{c|}{SCINet} & \multicolumn{2}{c|}{TiDE} & \multicolumn{2}{c}{DLinear} \\
\multicolumn{2}{c|}{\multirow{-2}{*}{Models}} & MSE                          & MAE                          & MSE                          & MAE                          & MSE                          & MAE                          & MSE                          & MAE   & MSE                          & MAE                          & MSE          & MAE         & MSE         & MAE        & MSE          & MAE          \\ \midrule
                                 & 96        & \textbf{{\color[HTML]{FE0000} 0.374}} & \textbf{{\color[HTML]{FE0000} 0.397}} & 0.386                        & 0.405                        & 0.414                        & 0.419                        & 0.376                        & 0.419 & 0.384                        & 0.402                        & 0.654        & 0.599       & 0.479       & 0.464      & 0.386        & 0.400        \\
                                 & 192       & \textbf{{\color[HTML]{FE0000} 0.419}} & \textbf{{\color[HTML]{FE0000} 0.429}} & 0.441                        & 0.436                        & 0.460                        & 0.445                        & 0.420                        & 0.448 & 0.436                        & \textbf{{\color[HTML]{FE0000} 0.429}} & 0.719        & 0.631       & 0.525       & 0.492      & 0.437        & 0.432        \\
                                 & 336       & 0.461                        & \textbf{{\color[HTML]{FE0000} 0.450}} & 0.487                        & 0.458                        & 0.501                        & 0.466                        & \textbf{{\color[HTML]{FE0000} 0.459}} & 0.465 & 0.491                        & 0.469                        & 0.778        & 0.659       & 0.565       & 0.515      & 0.481        & 0.459        \\
\multirow{-4}{*}{\rotatebox{90}{ETTh1}}          & 720       & \textbf{{\color[HTML]{FE0000} 0.474}} & \textbf{{\color[HTML]{FE0000} 0.467}} & 0.503                        & 0.491                        & 0.500                        & 0.488                        & 0.506                        & 0.507 & 0.521                        & 0.500                        & 0.836        & 0.699       & 0.594       & 0.558      & 0.519        & 0.516        \\ \midrule
                                 & 96        & 0.301                        & 0.353                        & \textbf{{\color[HTML]{FE0000} 0.297}} & \textbf{{\color[HTML]{FE0000} 0.349}} & 0.302                        & 0.348                        & 0.358                        & 0.397 & 0.340                        & 0.374                        & 0.707        & 0.621       & 0.400       & 0.440      & 0.333        & 0.387        \\
                                 & 192       & \textbf{{\color[HTML]{FE0000} 0.379}} & 0.405                        & 0.380                        & \textbf{{\color[HTML]{FE0000} 0.400}} & 0.388                        & \textbf{{\color[HTML]{FE0000} 0.400}} & 0.429                        & 0.439 & 0.402                        & 0.414                        & 0.860        & 0.689       & 0.528       & 0.509      & 0.477        & 0.476        \\
                                 & 336       & 0.432                        & 0.446                        & 0.428                        & \textbf{{\color[HTML]{FE0000} 0.432}} & \textbf{{\color[HTML]{FE0000} 0.426}} & 0.433                        & 0.496                        & 0.487 & 0.452                        & 0.452                        & 1.000        & 0.744       & 0.643       & 0.571      & 0.594        & 0.541        \\
\multirow{-4}{*}{\rotatebox{90}{ETTh2}}          & 720       & 0.446                        & 0.463                        & \textbf{{\color[HTML]{FE0000} 0.427}} & \textbf{{\color[HTML]{FE0000} 0.445}} & 0.431                        & 0.446                        & 0.463                        & 0.474 & 0.462                        & 0.468                        & 1.249        & 0.838       & 0.874       & 0.679      & 0.831        & 0.657        \\  \midrule
                                 & 96        & \textbf{{\color[HTML]{FE0000} 0.320}} & \textbf{{\color[HTML]{FE0000} 0.358}} & 0.334                        & 0.368                        & 0.329                        & 0.367                        & 0.379                        & 0.419 & 0.338                        & 0.375                        & 0.418        & 0.438       & 0.364       & 0.387      & 0.345        & 0.372        \\
                                 & 192       & \textbf{{\color[HTML]{FE0000} 0.364}} & \textbf{{\color[HTML]{FE0000} 0.383}} & 0.377                        & 0.391                        & 0.367                        & 0.385                        & 0.426                        & 0.441 & 0.374                        & 0.387                        & 0.439        & 0.450       & 0.398       & 0.404      & 0.380        & 0.389        \\
                                 & 336       & \textbf{{\color[HTML]{FE0000} 0.395}} & \textbf{{\color[HTML]{FE0000} 0.405}} & 0.426                        & 0.420                        & 0.399                        & 0.410                        & 0.426                        & 0.441 & 0.410                        & 0.411                        & 0.490        & 0.485       & 0.428       & 0.425      & 0.413        & 0.413        \\
\multirow{-4}{*}{\rotatebox{90}{ETTm1}}          & 720       & 0.457                        & 0.440                        & 0.491                        & 0.459                        & \textbf{{\color[HTML]{FE0000} 0.454}} & \textbf{{\color[HTML]{FE0000} 0.439}} & 0.543                        & 0.490 & 0.478                        & 0.450                        & 0.595        & 0.550       & 0.487       & 0.461      & 0.474        & 0.453        \\ \midrule
                                 & 96        & 0.176                        & 0.261                        & 0.180                        & 0.264                        & \textbf{{\color[HTML]{FE0000} 0.175}} & \textbf{{\color[HTML]{FE0000} 0.259}} & 0.203                        & 0.287 & 0.187                        & 0.267                        & 0.286        & 0.377       & 0.207       & 0.305      & 0.193        & 0.292        \\
                                 & 192       & \textbf{{\color[HTML]{FE0000} 0.240}} & \textbf{{\color[HTML]{FE0000} 0.302}} & 0.250                        & 0.309                        & 0.241                        & 0.302                        & 0.269                        & 0.328 & 0.249                        & 0.309                        & 0.399        & 0.445       & 0.290       & 0.364      & 0.284        & 0.362        \\
                                 & 336       & \textbf{{\color[HTML]{FE0000} 0.299}} & \textbf{{\color[HTML]{FE0000} 0.342}} & 0.311                        & 0.348                        & 0.305                        & 0.343                        & 0.325                        & 0.366 & 0.321                        & 0.351                        & 0.637        & 0.591       & 0.377       & 0.422      & 0.369        & 0.427        \\
\multirow{-4}{*}{\rotatebox{90}{ETTm2}}          & 720       & \textbf{{\color[HTML]{FE0000} 0.397}} & 0.401                        & 0.412                        & 0.407                        & 0.402                        & \textbf{{\color[HTML]{FE0000} 0.400}} & 0.421                        & 0.415 & 0.408                        & 0.403                        & 0.960        & 0.735       & 0.558       & 0.524      & 0.554        & 0.522        \\ \midrule
                                 & 96        & 0.166                        & 0.256                        & \textbf{{\color[HTML]{FE0000} 0.148}} & \textbf{{\color[HTML]{FE0000} 0.240}} & 0.181                        & 0.270                        & 0.193                        & 0.308 & 0.168                        & 0.272                        & 0.247        & 0.345       & 0.237       & 0.329      & 0.197        & 0.282        \\
                                 & 192       & 0.187                        & 0.274                        & \textbf{{\color[HTML]{FE0000} 0.162}} & \textbf{{\color[HTML]{FE0000} 0.253}} & 0.188                        & 0.274                        & 0.201                        & 0.315 & 0.184                        & 0.289                        & 0.257        & 0.355       & 0.236       & 0.330      & 0.196        & 0.285        \\
                                 & 336       & 0.204                        & 0.290                        & \textbf{{\color[HTML]{FE0000} 0.178}} & \textbf{{\color[HTML]{FE0000} 0.269}} & 0.204                        & 0.293                        & 0.214                        & 0.329 & 0.198                        & 0.300                        & 0.269        & 0.369       & 0.249       & 0.344      & 0.209        & 0.301        \\
\multirow{-4}{*}{\rotatebox{90}{ECL}}            & 720       & 0.247                        & 0.323                        & 0.225                        & \textbf{{\color[HTML]{FE0000} 0.317}} & 0.246                        & 0.324                        & 0.246                        & 0.355 & \textbf{{\color[HTML]{FE0000} 0.220}} & 0.320                        & 0.299        & 0.390       & 0.284       & 0.373      & 0.245        & 0.333        \\ \midrule
                                 & 96        & 0.511                        & 0.324                        & \textbf{{\color[HTML]{FE0000} 0.395}} & \textbf{{\color[HTML]{FE0000} 0.268}} & 0.462                        & 0.295                        & 0.587                        & 0.366 & 0.593                        & 0.321                        & 0.788        & 0.499       & 0.805       & 0.493      & 0.650        & 0.396        \\
                                 & 192       & 0.529                        & 0.330                        & \textbf{{\color[HTML]{FE0000} 0.417}} & \textbf{{\color[HTML]{FE0000} 0.276}} & 0.466                        & 0.296                        & 0.604                        & 0.373 & 0.617                        & 0.336                        & 0.789        & 0.505       & 0.756       & 0.474      & 0.598        & 0.370        \\
                                 & 336       & 0.545                        & 0.334                        & \textbf{{\color[HTML]{FE0000} 0.433}} & \textbf{{\color[HTML]{FE0000} 0.283}} & 0.482                        & 0.304                        & 0.621                        & 0.383 & 0.629                        & 0.336                        & 0.797        & 0.508       & 0.762       & 0.477      & 0.605        & 0.373        \\
\multirow{-4}{*}{\rotatebox{90}{Traffic}}        & 720       & 0.580                        & 0.351                        & \textbf{{\color[HTML]{FE0000} 0.467}} & \textbf{{\color[HTML]{FE0000} 0.302}} & 0.514                        & 0.322                        & 0.626                        & 0.382 & 0.640                        & 0.350                        & 0.841        & 0.523       & 0.719       & 0.449      & 0.645        & 0.394        \\ \midrule
                                 & 96        & \textbf{{\color[HTML]{FE0000} 0.164}} & \textbf{{\color[HTML]{FE0000} 0.210}} & 0.174                        & 0.214                        & 0.177                        & 0.218                        & 0.217                        & 0.296 & 0.172                        & 0.220                        & 0.221        & 0.306       & 0.202       & 0.261      & 0.196        & 0.255        \\
                                 & 192       & \textbf{{\color[HTML]{FE0000} 0.210}} & \textbf{{\color[HTML]{FE0000} 0.251}} & 0.221                        & 0.254                        & 0.225                        & 0.259                        & 0.276                        & 0.336 & 0.219                        & 0.261                        & 0.261        & 0.340       & 0.242       & 0.298      & 0.237        & 0.296        \\
                                 & 336       & \textbf{{\color[HTML]{FE0000} 0.265}} & \textbf{{\color[HTML]{FE0000} 0.290}} & 0.278                        & 0.296                        & 0.278                        & 0.297                        & 0.339                        & 0.380 & 0.280                        & 0.306                        & 0.309        & 0.378       & 0.287       & 0.335      & 0.283        & 0.335        \\
\multirow{-4}{*}{\rotatebox{90}{Weather}}        & 720       & \textbf{{\color[HTML]{FE0000} 0.343}} & \textbf{{\color[HTML]{FE0000} 0.342}} & 0.358                        & 0.349                        & 0.354                        & 0.348                        & 0.403                        & 0.428 & 0.365                        & 0.359                        & 0.377        & 0.427       & 0.351       & 0.386      & 0.345        & 0.381        \\  \midrule
\multicolumn{2}{c|}{Rank 1}                   & \multicolumn{2}{c|}{\textbf{{\color[HTML]{FE0000} 27}}}               & \multicolumn{2}{c|}{21}                                      & \multicolumn{2}{c|}{8}                                       & \multicolumn{2}{c|}{1}                & \multicolumn{2}{c|}{2}                                       & \multicolumn{2}{c|}{0}      & \multicolumn{2}{c|}{0}    & \multicolumn{2}{c}{0}      \\ \bottomrule
\end{tabular}
\caption{Multivariate time series forecasting results. Forecasting horizons $P$ is {96, 192, 336, 720}, and input length $T$ is set as 96. A lower value indicates better performance. The best results are marked in {\color[HTML]{FE0000}red}.} 
\label{tab:overall}
\end{table*}

\section{Experiments}
In this section, we experimentally validate the effectiveness of KAN on seven public benchmarks. Specifically, we conduct extensive experiments, including performance comparison, running speed comparison, integrating the KANs into other advanced models, and the interpretability of MMK.


\begin{table*}[!]
\centering
\begin{tabular}{cc|cccc|cccccc}
\midrule
                          &                     & \multicolumn{4}{c|}{with MoE}                                                                                                                                  & \multicolumn{6}{c}{w/o MoE}                                                                                                                                                                                            \\ \cmidrule{3-12}
                          &                     & \multicolumn{2}{c}{MoK}                                                    & \multicolumn{2}{c|}{MoL}                                                    & \multicolumn{2}{c}{WavKAN}                                                     & \multicolumn{2}{c}{TaylorKAN}                              & \multicolumn{2}{c}{Linear}                                                   \\
\multirow{-3}{*}{Dataset} & \multirow{-3}{*}{P} & MSE                                   & MAE                                   & MSE                                   & MAE                                   & MSE                                   & MAE                                   & MSE            & MAE                                   & MSE                                   & MAE                                   \\ \midrule
                          & 96                  & {\color{red} \textbf{0.326}} & {\color{red} \textbf{0.360}} & 0.335                                 & 0.367                                 & \textbf{0.341}                        & \textbf{0.370}                        & 0.340          & 0.368                                 & 0.355                                 & 0.376                                 \\
                          & 192                 & {\color{red} \textbf{0.367}} & {\color{red} \textbf{0.382}} & 0.374                                 & 0.385                                 & \textbf{0.374}                        & \textbf{0.384}                        & 0.380          & 0.386                                 & 0.391                                 & 0.392                                 \\
                          & 336                 & {\color{red} \textbf{0.400}} & {\color{red} \textbf{0.404}} & 0.407                                 & 0.406                                 & \textbf{0.406}                        & \textbf{0.405}                        & 0.412          & 0.406                                 & 0.424                                 & 0.415                                 \\
                          & 720                 & {\color{red} \textbf{0.462}} & \textbf{0.439}                        & 0.470                                 & 0.440                                 & \textbf{0.465}                        & {\color{red} \textbf{0.438}} & 0.474          & 0.439                                 & 0.487                                 & 0.450                                 \\
\multirow{-5}{*}{ETTm1}  & avg                 & {\color{red} \textbf{0.389}} & {\color{red} \textbf{0.396}} & 0.397                                 & 0.400                                 & \textbf{0.397}                        & \textbf{0.399}                        & 0.402          & 0.400                                 & 0.414                                 & 0.407                                 \\ \midrule
                          & 96                  & {\color{red} \textbf{0.382}} & 0.396                                 & 0.382                                 & {\color{red} \textbf{0.392}} & 0.396                                 & 0.402                                 & 0.388          & 0.398                                 & \textbf{0.386}                        & \textbf{0.395}                        \\
                          & 192                 & {\color{red} \textbf{0.430}} & 0.426                                 & 0.436                                 & {\color{red} \textbf{0.423}} & 0.439                                 & 0.430                                 & 0.438          & \textbf{0.424}                        & \textbf{0.437}                        & \textbf{0.424}                        \\
                          & 336                 & {\color{red} \textbf{0.468}} & {\color{red} \textbf{0.443}} & 0.487                                 & 0.449                                 & 0.479                                 & 0.449                                 & \textbf{0.477} & \textbf{0.442}                        & 0.479                                 & 0.446                                 \\
                          & 720                 & {\color{red} \textbf{0.450}} & {\color{red} \textbf{0.458}} & 0.486                                 & 0.472                                 & \textbf{0.464}                        & \textbf{0.461}                        & 0.477          & \textbf{0.461}                        & 0.481                                 & 0.470                                 \\
\multirow{-5}{*}{ETTh1}   & avg                 & {\color{red} \textbf{0.433}} & {\color{red} \textbf{0.431}} & 0.448                                 & 0.434                                 & \textbf{0.444}                        & 0.435                                 & 0.445          & {\color{red} \textbf{0.431}} & 0.446                                 & 0.434                                 \\ \midrule
                          & 96                  & 0.175                                 & 0.225                                 & {\color{red} \textbf{0.174}} & {\color{red} \textbf{0.223}} & \textbf{0.180}                        & \textbf{0.228}                        & 0.182          & 0.232                                 & 0.192                                 & 0.232                                 \\
                          & 192                 & {\color{red} \textbf{0.224}} & {\color{red} \textbf{0.267}} & 0.226                                 & 0.267                                 & \textbf{0.227}                        & \textbf{0.268}                        & 0.235          & 0.274                                 & 0.240                                 & 0.271                                 \\
                          & 336                 & {\color{red} \textbf{0.284}} & {\color{red} \textbf{0.308}} & 0.286                                 & {\color{red} \textbf{0.308}} & {\color{red} \textbf{0.284}} & {\color{red} \textbf{0.308}} & 0.291          & 0.312                                 & 0.292                                 & 0.307                                 \\
                          & 720                 & \textbf{0.366}                        & \textbf{0.358}                        & 0.367                                 & 0.358                                 & 0.368                                 & 0.360                                 & 0.370          & 0.360                                 & {\color{red} \textbf{0.364}} & {\color{red} \textbf{0.353}} \\
\multirow{-5}{*}{Weather} & avg                 & {\color{red} \textbf{0.262}} & 0.290                                 & 0.263                                 & {\color{red} \textbf{0.289}} & \textbf{0.264}                        & \textbf{0.291}                        & 0.269          & 0.294                                 & 0.272                                 & \textbf{0.291}   \\ \bottomrule                      
\end{tabular}
\caption{The performance of KAN-based models vs. Linear-based models. Both the best results of models with or w/o MoE are highlighted in bold, and the best models of all methods are marked in red.}
\label{tab:kan-vs-liner}
\end{table*}

\subsection{Experimental Settings}
\paragraph{Datasets} We conduct extensive experiments on seven widely-used datasets, including ETT(h1, h2, m1, m2) \cite{zhou2021informer}, ECL, Traffic and Weather \cite{lai2018modeling}, whose statistical information is shown in Table \ref{tab:dataset}. We follow the same data processing operations used in TimesNet \cite{wu2023timesnet}, where the training, validation, and testing sets are divided according to chronological order.

\paragraph{Settings} We implement our method and baselines in a unify code library with PyTorch \footnote{https://github.com/pytorch/pytorch} and Pytorch-lightning \footnote{https://github.com/Lightning-AI/pytorch-lightning} , and we conduct all experiments on a GPU server with NVIDIA A100 80GB GPUs. In order to prevent the `Drop Last' \cite{qiu2024tfb} in the test stage where the samples of the last incomplete batch are discarded, we set the test batch size to 1. We use the Mean Squared Error (MSE) and the Mean Absolute Error (MAE) as metrics. We use KAN, TaylorKAN, JacobiKAN and WavKAN to construct the MoK layer. The code in the supplementary files provides more details.

\paragraph{Baselines} We select seven well-known baselines, including (1) Transformer-based methods: iTransformer \cite{liu2024itransformer}, PatchTST \cite{nie2023patchtst} and FEDformer \cite{zhou2022fedformer}; (2) CNN-based methods: TimesNet \cite{wu2023timesnet} and SCINet \cite{liu2022SCINet}; (3) Linear-based methods: TiDE \cite{das2023tide} and DLinear \cite{zeng2023dlinear}.

\subsection{Can KANs Get the SOTA Performance?}
\label{sec:exp-1}

In this section, we compare our proposed MMK with 7 popular Transformer-based, CNN-based and Linear-based baselines on seven custom datasets to comprehensively evaluate the performance of KAN. The experimental results are shown in Table \ref{tab:overall}, where the best results of all models are marked in red and the second best results are marked with underlines. The results of baselines come from previous paper \cite{liu2024itransformer}, and the results of MMK are average of four times experiments with fixed seed [0, 1, 2, 3].
Surprisingly, MMK achieves the best performance in most cases. And it significantly outperforms on the Weather dataset, reducing the MSE by 5.74\% on the 96-step forecasting compared to the second-best method, iTransformer. We infer that this improvement comes from the fact that KAN's function-fitting ability is more suitable for processing natural science data. The performance on traffic data is poor because our method does not design a module to exploit the correlation between variables. Subsequent research can consider designing a KAN-based variable relationship mining module to better improve model performance.
In general, a properly designed \textbf{\uline{KAN-based model can achieve the state-of-the-art performance}} in most cases while maintaining the advantages of interpretability. 

\begin{table}[t]
\small
\centering
\begin{tabular}{cc|cc|cc|cc}
\midrule
\multicolumn{2}{c|}{\multirow{2}{*}{Dataset}} & \multicolumn{2}{c|}{iTransformer} & \multicolumn{2}{c|}{+KAN}  & \multicolumn{2}{c}{+MoK} \\
& & MSE & MAE & MSE & MAE  & MSE & MAE \\ \midrule
\multirow{5}{*}{\rotatebox{90}{ETTh1}} & 96 & \textbf{{\color[HTML]{FE0000}0.390}} & \textbf{{\color[HTML]{FE0000}0.404}}  & 0.397 & 0.409 & 0.393 & 0.409 \\ 
 & 192 & 0.445 & \textbf{{\color[HTML]{FE0000}0.436}} & 0.443 & 0.436  & \textbf{{\color[HTML]{FE0000}0.442}} & 0.441 \\
 & 336 & 0.489 & \textbf{{\color[HTML]{FE0000}0.457}} & 0.491 & 0.461  & \textbf{{\color[HTML]{FE0000}0.476}} & 0.458 \\
 & 720 & 0.505 & 0.487 & 0.510 & 0.488  & \textbf{{\color[HTML]{FE0000}0.494}} & \textbf{{\color[HTML]{FE0000}0.485}} \\
 & avg & 0.457 & \textbf{{\color[HTML]{FE0000}0.446}} & 0.460 & 0.448  & \textbf{{\color[HTML]{FE0000}0.451}} & 0.448 \\ \midrule
\multirow{5}{*}{\rotatebox{90}{ETTm1}} & 96 & 0.341 & 0.373 & 0.331 & 0.365  & \textbf{{\color[HTML]{FE0000}0.329}} & \textbf{{\color[HTML]{FE0000}0.364}} \\
 & 192 & 0.388 & 0.397 & 0.379 & 0.392 & \textbf{{\color[HTML]{FE0000}0.376}} & \textbf{{\color[HTML]{FE0000}0.390}} \\
 & 336 & 0.426 & 0.420 & 0.417 & 0.416  & \textbf{{\color[HTML]{FE0000}0.412}} & \textbf{{\color[HTML]{FE0000}0.415}} \\
 & 720 & 0.498 & 0.458 & 0.484 & \textbf{{\color[HTML]{FE0000}0.450}}  & \textbf{{\color[HTML]{FE0000}0.478}} & 0.453 \\
 & avg & 0.413 & 0.412 & 0.403 & \textbf{{\color[HTML]{FE0000}0.406}}  & \textbf{{\color[HTML]{FE0000}0.399}} & \textbf{{\color[HTML]{FE0000}0.406}} \\ 
 \bottomrule 
\end{tabular}
\caption{The performance of KAN within iTransformer. The best results of all methods are highlighted in red.}
\label{tab:kan-in-transformer}
\end{table}

\subsection{Does KANs Outperform Linear?}
\label{sec:exp-2}
In this section, we conduct ablation experiments to compare KAN with Linear in time series forecasting tasks. For fair comprison, we only use one layer of forward network that means all models in this experiment only consist of RevIN and a single network layer.
\begin{itemize}
    \item MoK: It uses four KAN variants (KAN, WavKAN, TaylorKAN, and JacobiKAN) as experts for forecasting.
    \item MoL: It uses four Linears as experts for forecasting.
    \item WavKAN, TaylorKAN, Linear: They use the corresponding layer for forecasting.
\end{itemize}

The experimental results are reported in Table \ref{tab:kan-vs-liner}, where all the results are the average of 4 times experiments, and the best results are highlighted in red. 
We can conclude three useful empirical experiences.
First, the KAN-based model outperforms the linear-based model in most cases. We speculate this phenomenon is due to KAN's function representation idea, which is more efficient to capture the periodicity and trend in time series.
Second, the mixture of expert structures is applicable to both KAN and Linear, which should be attributed to the fact that the gating network assigns variables to specific experts.
Third, the performance of KAN-based models is affected by the specific function, which may be related to the intrinsic distribution of the time series. Through these two groups of experiments, we can conclude that \textbf{\uline{KAN has competitive performance compared to Linear}} in time-series forecasting.

\begin{table}[t]
\centering
\begin{tabular}{c|cc|cc}
\toprule
\multirow{2}{*}{Models} & \multicolumn{2}{c|}{ETTh1} & \multicolumn{2}{c}{ETTm1} \\
                        & MSE         & MAE         & MSE         & MAE         \\ \midrule
TimeMixer           & 0.375       & 0.400       & 0.320       & 0.357       \\
+MoK                    & 0.375       & 0.398       & 0.314       & 0.353       \\
Improve                 & 0.00\%      & 0.50\%      & 1.88\%      & 1.12\%      \\ \midrule
GBT                 & 0.398       & 0.418       & 0.319       & 0.370       \\
+MoK                    & 0.396       & 0.417       & 0.311       & 0.370       \\
Improve                 & 0.50\%      & 0.24\%      & 2.51\%      & 0.00\%      \\ \midrule
Pathformer          & 0.382       & 0.400       & 0.316       & 0.346       \\
+MoK                    & 0.379       & 0.389       & 0.315       & 0.343       \\
Improve                 & 0.79\%      & 2.75\%      & 0.32\%      & 0.87\%      \\ \bottomrule
\end{tabular}
\caption{The performance of KAN within other model.}
\label{tab:kan-in-more-model}
\end{table}

\subsection{Can KANs be Integrated into Other Models?}
\label{sec:exp-3}
In Section \ref{sec:exp-1}, we have demonstrated that our proposed MMK has excellent performance compared to models based on other network architectures. In order to further explore the upper bound of KAN’s performance in time series forecasting, we integrate KAN into some existing time series forecasting models in the form of replacing the original predictor so as to make full use of the progress of advanced time series forecasting technology.
First, we select iTransformer \cite{liu2024itransformer} as baseline and replace the last linear predictor with a KAN or MoK layer. For fair comparison, we set the same network hyperparameters in all experiments where hidden dim is 512 and layer number is 2, and search the best learning rate from 1e-2 to 1e-5 through a grid searching strategy. As shown in Table \ref{tab:kan-in-transformer}, iTransformer with KAN and MoK can achieve better performance in most cases. Then, we intergate MoK with more models \cite{wang2024timemixer,chen2024pathformer,shen2023gbt} in Table \ref{tab:kan-in-more-model}, MoK can achieve better performance in most cases. These experimental findings demonstrate that \textbf{\uline{MoK is a powerful form for integrating KAN into existing models}}.

\subsection{How to Get Better MMK?}
\label{sec:exp-4}
\begin{figure}[t]
\centering
\includegraphics[width=0.4\textwidth]{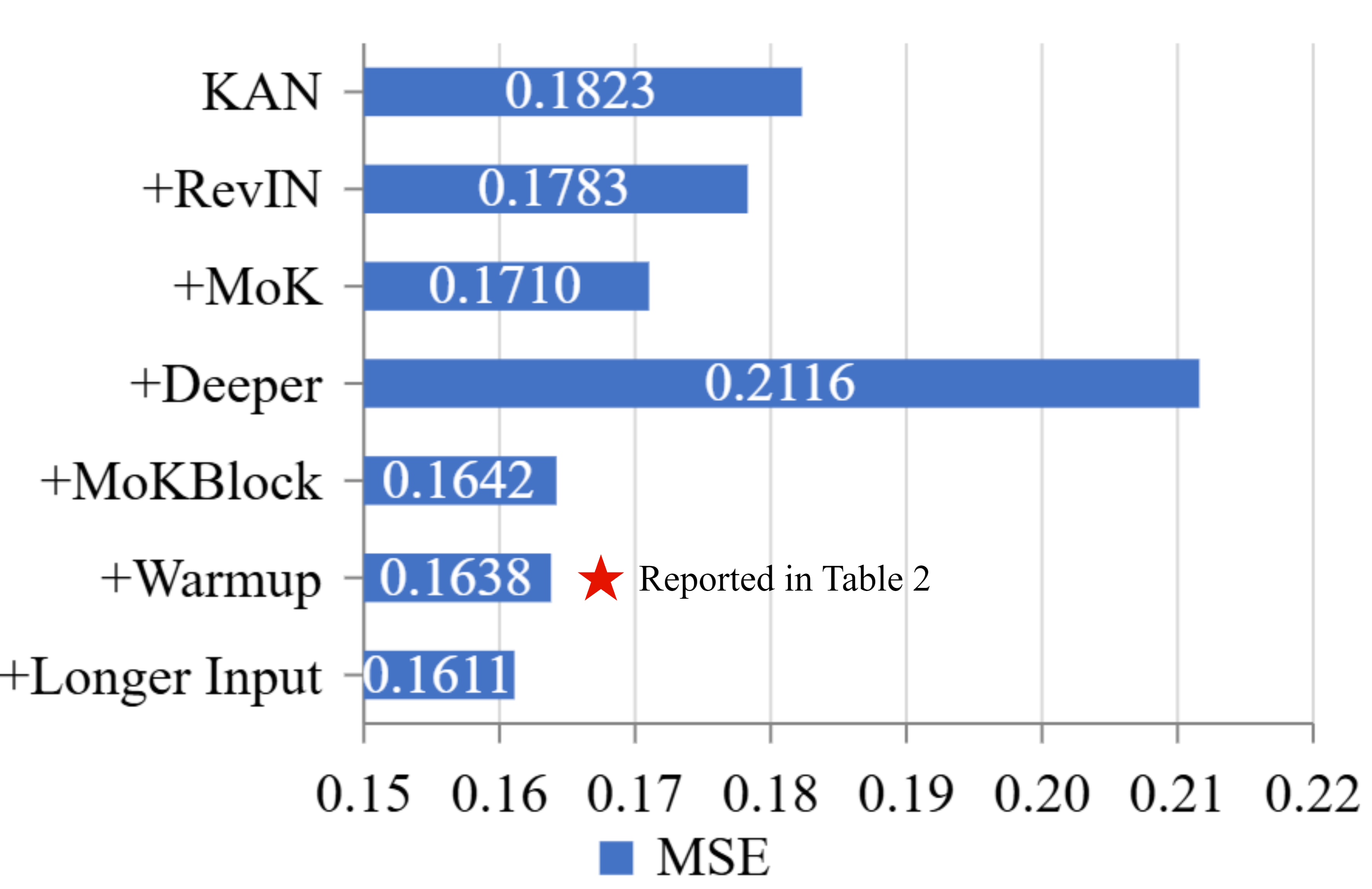}
\caption{The impact of different modules or training strategies.}
\label{fig:tricks}
\end{figure}

We evaluate the impact of different modules or training strategies on the forecasting performance. The pre-sampling weights initialization strategy is used in all experiments to maintain the stability of training. As shown in Figure \ref{fig:tricks}, we report the MSE of $L$=96 and $P$=96, except for `+Longer Input' where $L$=192 and $P$=96. First, we can find that MoK and RevIN are useful to enhance the forecasting ability. Then, adding model depth (`+MoK') only will drop the performance significance. The MoKBlock module and the warmup strategy can improve the performance of multi-layer MMK to achieve the best results. We also report the result under `+Longer Input' to illustrate that MMK can get enhancement from longer input. 

\subsection{Are KANs Efficiency?}
\label{sec:exp-5}
We report the size of model parameters, the training and inferring speed of KAN-based methods and baselines on ETTh1 dataset with input length 96 and prediction length 720. We set training batch size to 64, infer batch size to 1. As shown in Table \ref{tab:efficiency}, MMK has more parameters than DLinear, but significantly less than transformer-based methods like iTransformer and PatchTST. The running speed of MMK is slower than that of DLinear. We believe that as researchers continue to optimize the computational efficiency of KAN, the running speed of MMK will be improved in the future.

\begin{table}[]
\centering 
\begin{tabular}{c|ccc}
\midrule
Model        & Param  & Training (it/s) & Infer (it/s) \\ \midrule
DLinear      & 139 K  & 102.51 & 360.93 \\
iTransformer & 3.6 M  & 51.03  & 188.79 \\
PatchTST     & 7.6 M  & 43.92  & 193.58 \\ \midrule
MMK          & 972 K  & 34.86 & 190.57 \\ \bottomrule 
\end{tabular}
\caption{Comparison of model parameters and running speed of models. All results are tested with $L$=96 and $P$=720 on ETTh1. }
\label{tab:efficiency}
\end{table}


\subsection{Case Study}
\label{sec:exp-7}

\begin{figure}[t]
\centering
\includegraphics[width=0.4\textwidth]{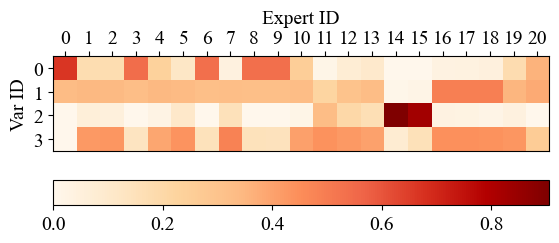}
\caption{Heatmap of the expert loads in the MoK on the Weather dataset. The value at location \textit{xy} equal to 1 means that the gating network assigns the \textit{y}-th variable to the \textit{x}-th expert in all samples.}
\label{fig:score}
\end{figure}

\begin{figure}[t]
\centering
\includegraphics[width=0.4\textwidth]{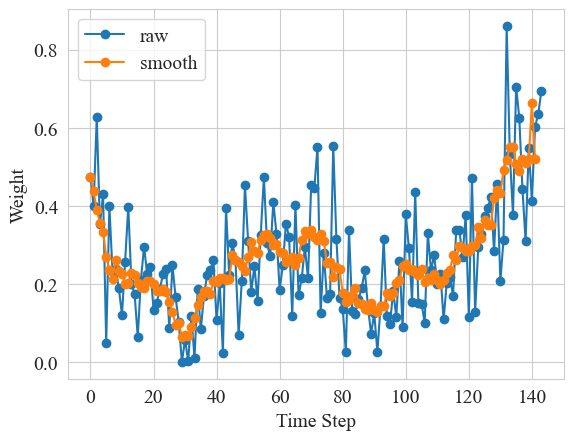}
\caption{The weight of each input feature.}
\label{fig:subfig2}
\end{figure}

In this section, we aim to analyze the interpretability of MMK for the time series forecasting task. First, we generate a heatmap to visualize the outputs of the gating network, which decide variable-expert assignment. The gating network outputs the matching score between variables and experts based on the input feature, and the top-$k$ matching experts are selected for each variable to predict the future state. 
We count the top-1 scores of all samples in the weather dataset and generate a heatmap in Figure \ref{fig:score}, where (x, y) = 1.0 means the \textit{y}-th variable matches the \textit{x}-th expert in all samples. 
We can see that the variable is closely related to the specific expert, which indicates that MoK can effectively assign KANs to variables.

Then, we use the temperature series in the weather dataset, which collects every 10 minutes to train the KAN model. We input one daily period (the past 144 time steps) of data to predict the state of the next time step. We visualize the weight of each feature dimension of the model. As shown in Figure \ref{fig:subfig2}, there are three peaks in the feature weight, which are near 0, 72, and 144 time steps. 0 represents the temperature at the same time step of the previous day's period, 144 represents the temperature at the adjacent moment, and 70 represents half a period. We conclude that \textbf{\uline{KAN can learn the periodicity of time series}}, which can preliminary explain its effectiveness in time series prediction.


\section{Conclusion}

The Kolmogorov-Arnold Network (KAN) seems to provide a new perspective for interpretable time series forecasting. In this paper, we evaluate the effectiveness of KANs in time-series forecasting tasks for the first time. First, we design the mixture-of-KAN layer, which uses a mixture-of-experts structure to assign variables to best-matched KAN experts. This module aggregates the advantages of various KAN's variants. Then, we propose the multi-layer mixture-of-KAN network (MMK), which achieves excellent performance while retaining KAN's ability to be transformed into a combination of symbolic functions. And we use some useful tricks to address the issues in the training stage. Finally, we compare MMK with many popular baselines on seven datasets. The effectiveness of KANs in multivariate time series forecasting has been proven by extensive experiments and visualizations. This work is only a preliminary attempt to introduce KAN into time series forecasting. We hope that our work can help future studies improve the performance and interpretability of KAN-based models.

\bibliographystyle{named}
\bibliography{ijcai25}

\end{document}